\documentclass{article}

\usepackage[numbers,sort]{natbib} 

\usepackage[pdftex]{graphicx}
\usepackage{makecell}
\usepackage{booktabs}     
\usepackage{amsmath}
\usepackage{colortbl}
\usepackage{xcolor}
\usepackage{siunitx}      
\usepackage{hyperref}
\usepackage{cleveref}
\usepackage{subcaption}
\usepackage[normalem]{ulem}
\usepackage{tabularx} 
\usepackage{enumitem}
\usepackage{multirow}
\usepackage{textcomp}  
\usepackage{scalerel}  
\usepackage{diagbox} 

\def\crossemoji{\scalerel*{\includegraphics{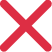}}{\textrm{\textbigcircle}}}
\def\checkemoji{\scalerel*{\includegraphics{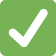}}{\textrm{\textbigcircle}}}

\newcommand{\datasetname}{DOCCI-Critique}
\newcommand{\modelname}{VNLI-Critique}

\definecolor{gray1}{gray}{0.95}
\definecolor{gray2}{gray}{0.91}
\definecolor{gray3}{gray}{0.87}
\definecolor{gray4}{gray}{0.83}
\definecolor{gray5}{gray}{0.79}
\definecolor{gray6}{gray}{0.75}
\definecolor{gray7}{gray}{0.71}
\definecolor{gray8}{gray}{0.67}
\definecolor{gray9}{gray}{0.63}
\definecolor{gray10}{gray}{0.59}
\definecolor{gray11}{gray}{0.55}
\definecolor{gray12}{gray}{0.51}
\definecolor{gray13}{gray}{0.47}
\definecolor{gray14}{gray}{0.43}

\newcommand{\rankcell}[2]{
    \cellcolor{gray#1}#1$^{\num[round-mode=places, round-precision=2]{#2}}$%
}

    \usepackage[preprint]{neurips_2025}

\usepackage[utf8]{inputenc} 
\usepackage[T1]{fontenc}    
\usepackage{hyperref}       
\usepackage{url}            
\usepackage{booktabs}       
\usepackage{amsfonts}       
\usepackage{nicefrac}       
\usepackage{microtype}      
\usepackage{xcolor}         

\title{Unblocking Fine-Grained Evaluation of Detailed Captions: An Explaining AutoRater and Critic-and-Revise Pipeline}

\author{Brian Gordon*$^{1,2}$, Yonatan Bitton*$^{2}$, Andreea Marzoca$^{2}$, 
Yasumasa Onoe$^{2}$, \\ \textbf{Xiao Wang}$^{2}$, \textbf{Daniel Cohen-Or}$^{1}$, \textbf{Idan Szpektor}$^{2}$, \\
       $^{1}$Tel Aviv University, $^{2}$Google Research \\
       {\color{cyan}\url{https://google.github.io/unblocking-detail-caption}}
        }
        
\begin{document}

\let\oldthefootnote\thefootnote
\let\thefootnote\relax\footnotetext{* Equal contribution}
\let\thefootnote\oldthefootnote

\maketitle

\begin{figure*}[h]
  \centering
\includegraphics[width=\textwidth]{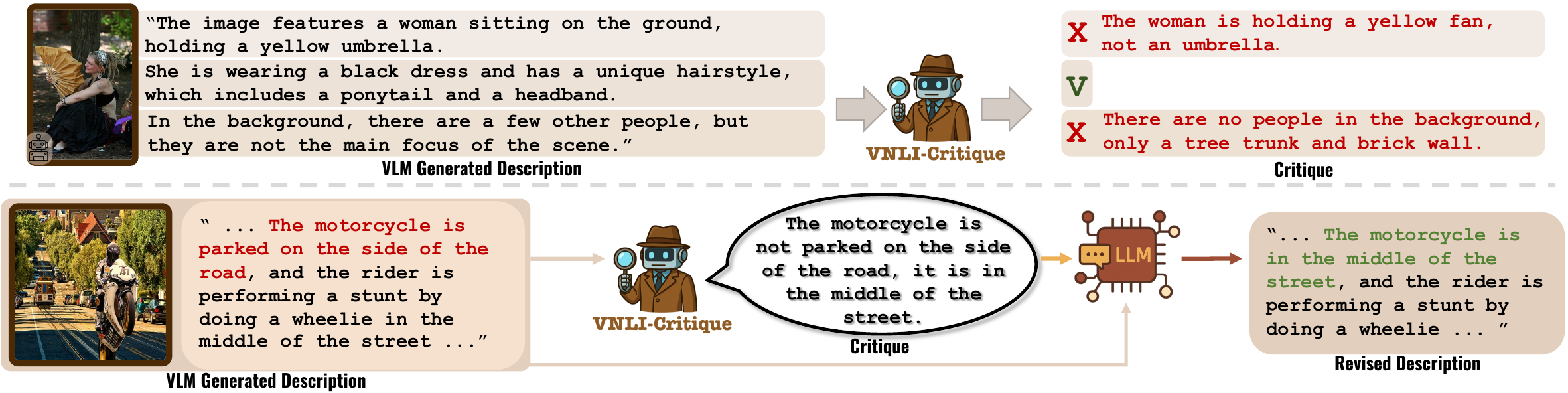}

\caption{\modelname{} in action: operating as a Critic and within the Critic-and-Revise pipeline. (Top) As a Critic, \modelname{} evaluates sentence-level factuality in VLM captions and generates error explanations. (Bottom) In the pipeline, its critique of an incorrect sentence guides an LLM to revise it, demonstrating automated evaluation and correction of detailed captions.}

\label{fig:teaser}

\end{figure*}
\begin{abstract}
Large Vision-Language Models (VLMs) now generate highly detailed, paragraph-length image captions, yet evaluating their factual accuracy remains challenging. Current methods often miss fine-grained errors, being designed for shorter texts or lacking datasets with verified inaccuracies. We introduce \textit{{\datasetname}}, a benchmark with 1,400 VLM-generated paragraph captions (100 images, 14 VLMs) featuring over 10,216 sentence-level human annotations of factual correctness and explanatory rationales for errors, all within paragraph context. Building on this, we develop \textit{\modelname{}}, a model for automated sentence-level factuality classification and critique generation. We highlight three key applications:
(1) \modelname{} demonstrates robust generalization, validated by state-of-the-art performance on the M-HalDetect benchmark and strong results in CHOCOLATE claim verification.
(2) The \modelname{} driven AutoRater for \datasetname{} provides reliable VLM rankings, showing excellent alignment with human factuality judgments (e.g., 0.98 Spearman).
(3) An innovative Critic-and-Revise pipeline, where critiques from \modelname{} guide LLM-based corrections, achieves substantial improvements in caption factuality (e.g., a 46\% gain on DetailCaps-4870).
Our work offers a crucial benchmark alongside practical tools, designed to significantly elevate the standards for fine-grained evaluation and foster the improvement of VLM image understanding.
\end{abstract}
\newpage
\begin{figure*}[t]
  \centering
  \includegraphics[width=\columnwidth]{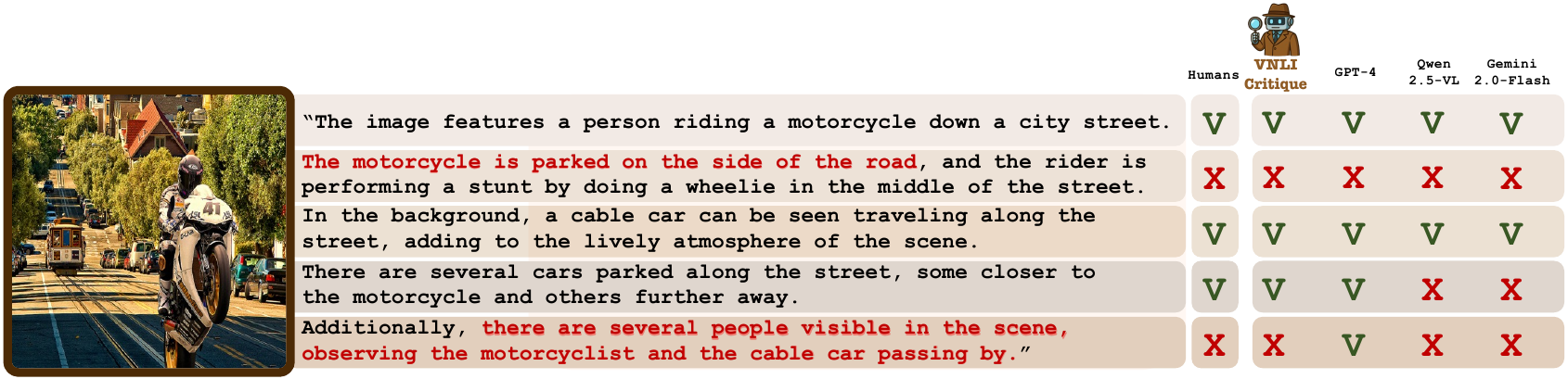}

\caption{Sentence-level factuality assessment by \modelname{}. Figure shows an image, VLM-generated caption, and factuality judgments ($V$ Correct / $X$ Incorrect), illustrating \modelname{}'s fine-grained, human-aligned fact-checking compared to zero-shot VLMs. Errors highlighted in red.}
\label{fig:competitors}
\end{figure*}
\section{Introduction}
\label{sec:intro}
Automatic descriptive image captioning, a prominent vision-language research area~\cite{survey_image_caption}, has evolved from short highlights~\cite{evolution_caption_models} to detailed, paragraph-length descriptions, thanks to powerful Large Vision-Language Models (LVLMs)~\cite{januspro, instructblip, llava_onevison, paligemma2,wang2024emu3, ye2024mplugowl3longimagesequenceunderstanding}. Evaluating these complex captions remains challenging; current metrics, often for short texts, miss subtle, fine-grained details and typically assess sentences in isolation, lacking crucial paragraph context for resolving ambiguities and co-references. While some studies address full paragraphs~\cite{CAPTURE}, granular sentence-level assessment is still difficult.

Evaluating VLM-generated text factuality has led to specialized benchmarks. However, many existing error/hallucination detection benchmarks target short sentences or QA tasks, inadequately addressing paragraph-length descriptions. While datasets like M-HalDetect~\cite{mhaldetect} and CHOCOLATE~\cite{CHOCOLATE} offer valuable sentence-level factuality annotations, they may lack full error diversity from long-form VLM outputs or consistent, detailed textual explanations for inaccuracies—vital for developing effective, fine-grained automated evaluation. Response-level evaluations (e.g., CAPTURE~\cite{CAPTURE}) score entire paragraphs but miss sentence-level granularity. Thus, a benchmark with comprehensive, context-aware sentence-level factuality annotations, including explanatory rationales for errors across diverse VLM-generated paragraphs, is critically needed.

To address this, we introduce \textit{\datasetname}, a novel benchmark for fine-grained evaluation of detailed image descriptions. It comprises 1,400 paragraph-length captions (14 VLMs, 100 images), with its core value in 10,216 sentence-level human annotations. Each sentence's factuality was judged by five annotators, with detailed textual rationales for every identified error. This multi-perspective annotation offers a new resource for in-depth VLM analysis, providing multiply-verified sentence-level judgments and error explanations for long captions, distinct from existing datasets.

Building on \datasetname{}, we developed \modelname{}, a model for automated sentence-level factuality classification (using paragraph context) and explanatory critique generation. This dual capability enables a novel Critic-and-Revise pipeline (Fig.~\ref{fig:teaser}): \modelname{} evaluates and generates a critique for an incorrect VLM-generated sentence, guiding an LLM to revise it. The utility of \modelname{} and this pipeline is shown via key results:
(1) \modelname{} achieves state-of-the-art performance on external benchmarks M-HalDetect (0.76 Macro-F1) and competitive results on CHOCOLATE~\cite{CHOCOLATE}, demonstrating strong generalization.
(2) On our benchmark, the \modelname{} powered \datasetname{} AutoRater shows VLM rankings with strong correlation to human judgments (\Cref{table:ranking_correlations}).
(3) The Critic-and-Revise pipeline significantly boosts factuality of incorrect sentences (e.g., by 46\% on DetailCaps-4870~\cite{CAPTURE}, 51\% on PixelProse~\cite{singla2024pixelsproselargedataset}), confirmed by human evaluation.
Collectively, these contributions deliver an essential benchmark and powerful methodologies, enabling more precise fine-grained assessment and demonstrably enhancing the factual accuracy of detailed image understanding by VLMs.
\section{Related Work}
\label{sec:related_work}
Our work intersects with advancements in Vision-Language Models (VLMs) for detailed captioning, the development of image captioning datasets, and methodologies for evaluating caption quality, especially factual accuracy and fine-grained detail.

\paragraph{Vision-Language Models}
Recent VLMs~\cite{paligemma2,openai2024gpt4ocard,Qwen2.5-VL,llava_onevison,LLAVA-1.5,wang2024emu3,MOLMO,mplug-owl2,chen2023pali,VILA,instructblip,zhu2023minigpt, geminiteam2024gemini15unlockingmultimodal} have achieved remarkable SOTA performance in multimodal tasks . Typically, they combine a visual encoder \cite{CLIP, SigLIP, SigLIP2, ViT} with an LLM \cite{qwen2_llm, FlanT5, vicuna2023, llama2}, often using a connector module to bridge visual features with textual tokens. Training usually involves pre-training the visual encoder and then fine-tuning the LLM, with objectives like masked language modeling or image-text matching. While end-to-end training is explored, this two-stage approach is common, balancing dataset scale, computational resources, and evaluation.

\paragraph{Image Captioning Datasets}
Image understanding and captioning datasets are crucial for supporting advancements in current image captioning techniques.  Early datasets provided positive image-text pairs with short sentences, primarily focusing on main objects and scenes, such as COCO \cite{COCO}, Flickr8k \cite{flickr8k}, and Flickr30k \cite{flickr30k}. More recent image captioning datasets offer image-text pairs with longer, more detailed descriptions. The Densely Captioned Images (DCI) dataset~\cite{DCI}, for instance, introduces long, mask-aligned descriptions to specifically evaluate VLM understanding of distinct image regions. PixelProse \cite{singla2024pixelsproselargedataset} offers a large scale dataset of 16.9 million synthetically generated captions using Gemini-1.0-Pro-Vision \cite{geminiteam2024gemini15unlockingmultimodal}; however, the correctness of these descriptions is not guaranteed.  The IIW~\cite{imageinwords} dataset utilizes a VLM to generate initial captions and then employs a human-in-the-loop framework to ensure high-quality positive image-text pairs.  M-HalDetect~\cite{mhaldetect} and CHOCOLATE \cite{CHOCOLATE} caption images and charts, respectively, using various VLMs, and employ sentence-level human annotation to assess the correctness of generated captions. DetailCaps~\cite{CAPTURE} leverages GPT-4~\cite{openai2024gpt4ocard} to assign a quality score (ranging from 0 to 5) to synthetically generated captions from existing datasets.  The DOCCI dataset \cite{DOCCI} stands out as a particularly strong choice for research due to its high-resolution images of diverse real-life scenes paired with detailed, fully human-written descriptions, offering a valuable resource for training and evaluating VLMs.  It provides high-resolution images of diverse real-life scenes, each paired with carefully crafted, long, human-annotated descriptions.

\paragraph{Evaluation of Detailed Captions}
Traditional metrics (e.g., BLEU~\cite{BLEU}, METEOR~\cite{banerjee-lavie-2005-meteor}, CIDEr~\cite{CIDER}) compare generated captions to references using n-gram overlap, often missing semantic nuance crucial for paragraph-length text. Embedding-based methods like CLIPScore~\cite{hessel-etal-2021-clipscore} and SIGLip~\cite{SigLIP} offer better semantic assessment but typically evaluate sentences in isolation, lacking the paragraph context needed to resolve ambiguities or co-references in detailed descriptions. QA-based metrics (e.g., VQAScore~\cite{vqascore}, TIFA~\cite{TIFA}, $\text{VQ}^2$~\cite{WYSIWYR}, GECKO~\cite{GECKO}) assess understanding via question-answering but face scalability challenges for long narratives. Recent work such as Mismatch Quest~\cite{mismatch_quest} has focused on providing detailed textual and visual feedback for misalignments, primarily in the text-to-image domain and often for shorter, direct textual inputs, by automatically generating explanations. While response-level evaluations like \citet{CAPTURE} score entire paragraphs, they don't offer sentence-level factuality details for descriptive captions. Our work addresses the need for fine-grained, context-aware sentence-level factuality evaluation with rich, human-verified explanatory feedback specifically for detailed, paragraph-length image captions
\section{The \datasetname{} Benchmark: Construction, Annotation, and Analysis}
\label{sec:docci100}

\textit{\datasetname} is a novel benchmark for fine-grained factuality assessment of paragraph-level image descriptions. Its primary purpose is twofold: (1) providing a robust platform to assess SOTA captioning models' descriptive capabilities and factual accuracy, and (2) serving as a challenging testbed for automated image understanding and fact-checking systems (auto-raters).

Construction began with 100 diverse, high-resolution images from the DOCCI dataset's `qual-dev' split~\cite{DOCCI}. For each, 14 SOTA Large Vision-Language Models (Table~\ref{tab:docci100_details}) generated detailed, paragraph-length descriptions, yielding 1,400 model-generated descriptions. This corpus deliberately captures a wide spectrum of stylistic variations, detail levels, and factual accuracies, from concise and factual to more verbose accounts that might introduce subtle inconsistencies (e.g., object misidentification).

The core of \datasetname{} is its rich human annotation layer. Following \citet{paligemma2} (see Appendix for annotation template), five human annotators independently evaluated each sentence within the 1,400 descriptions for factual correctness against the image, assigning labels: `Entailment' (factually supported), `Neutral' (not verifiable/contradicted), `Contradiction' (factually contradicted), or `Nothing to assess' (e.g., filler). A sentence is classified as `Non-entailment' if a majority labeled it `Neutral' or `Contradiction'. Crucially, annotators provided detailed textual rationales for each non-entailed judgment. Thus, a single non-entailed sentence often has multiple rationales, capturing diverse perspectives on the inaccuracy and offering insights into VLM error types (e.g., object misidentification, attribute/spatial errors, hallucination).

\begin{table*}[t]  
\caption{Illustrative example from the \datasetname{} benchmark, detailing sentence-level annotations for fine-grained factuality assessment, including rater judgments and explanatory rationales.} 
\centering
\scriptsize 
\setlength{\tabcolsep}{3pt} 
\renewcommand{\arraystretch}{0.9} 
\renewcommand\tabularxcolumn[1]{m{#1}}

\begin{tabularx}{\textwidth}{@{} *{4}{>{\centering\arraybackslash}X} @{}} 

\toprule
{\textbf{Image}} & \multicolumn{2}{c}{\raisebox{-0.5\height}{\includegraphics[width=0.25\textwidth]{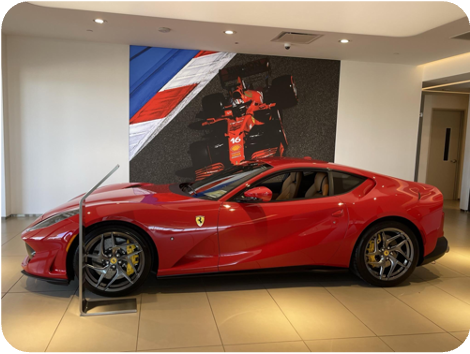}}} \\

\midrule
\textbf{\makecell{Description \\ Sentence}} & ``... Behind the car, there is a large mural or poster on the wall ...'' & ``... The mural features a Formula 1 racing car, also red, with the number 16 prominently displayed on the side. ...'' & ``... The background of the mural includes a racing track with the colors of the French flag (blue, white, and red) and a checkered flag, indicating a racing theme ...'' \\
\midrule
\textbf{Does the sentence include a claim about the image? (Answers from 5 raters)} 
    & \checkemoji, \checkemoji, \checkemoji, \checkemoji, \checkemoji 
    & \checkemoji, \checkemoji, \checkemoji, \checkemoji, \checkemoji 
    & \checkemoji, \checkemoji, \checkemoji, \checkemoji, \checkemoji \\
\midrule
\textbf{Is the sentence factual? (Answers from 5 raters)} 
    & \checkemoji, \checkemoji, \checkemoji, \checkemoji, \checkemoji 
    & \checkemoji, \checkemoji, \crossemoji, \crossemoji, \checkemoji 
    & \crossemoji, \checkemoji, \crossemoji, \crossemoji, \checkemoji \\
\midrule

\textbf{Rationales} & - & \begin{itemize}[leftmargin=*]
    \item The 16 number is not on the side of the racing car, but in the front of it.
    \item The mural does feature a red Formula 1 race car, but the number 16 is painted on the front not the side
\end{itemize} & \begin{itemize}[leftmargin=*] \item There is no checkered flag visible in the image. \item There's no checkered flag in the poster/mural. \item The background mural does feature the colors blue, white and red, but there is no checkered flag \end{itemize} \\
\bottomrule
\end{tabularx}
\label{table:annotation_example}
\vspace{-15pt}
\end{table*}

\Cref{table:annotation_example} illustrates this structure, showing per-sentence annotations: five independent factuality judgments (\checkemoji{} for Entailment, \crossemoji{} for Contradiction/Neutral) and image content reliance. For non-entailed votes, collected textual rationales explain the specific error, as exemplified by the car mural's number placement or the non-existent checkered flag. Multiple rationales for one incorrect sentence reflect diverse annotator viewpoints.

This comprehensive annotation resulted in 10,216 sentence-level judgments. The dataset—with its diverse VLM outputs, fine-grained majority-vote factuality labels, and multiple rich explanatory rationales per error—is an invaluable resource. It enables rigorous VLM evaluation beyond surface-level similarity, allowing deeper probes into model understanding and descriptive fidelity.

Table~\ref{tab:docci100_details} details per-model statistics from \datasetname, including description and sentence lengths, factual accuracy, and lexical diversity. These internal statistics reveal quantitative and qualitative VLM behavioral differences. For instance, high-accuracy models like GPT-4o contrast with Gemini models that produce more sentences with comparable accuracy, suggesting a verbosity/detail vs. error-risk trade-off. Paragraphs in \datasetname{} average 752.7 characters in length. This is considerably longer than in other contemporary datasets used for factual error analysis, such as M-HalDetect (averaging 456.2 characters), CHOCOLATE (577.6), and DetailCaps-4870 (612.9). This emphasis on longer, more intricate descriptions, combined with patterns like common error types (discernible from rationales, though not directly in Table~\ref{table:annotation_example}), highlights \datasetname's utility for nuanced comparative studies of VLM generation strategies and visual faithfulness, underscoring its role in evaluating diverse VLM behaviors in detailed captioning.

\begin{table*}[t]
\caption{\datasetname{} statistics, detailing paragraph-level metrics and lexical diversity (unique 2-grams) for each LLM's generated image descriptions.}
\centering
\resizebox{\textwidth}{!}{%
\begin{tabular}{l|cccccc}
\toprule
 & \makecell{Description\\Length avg.} & \makecell{\# Sentences\\Avg} & \makecell{Sentence\\Length Avg} & \makecell{\% Correct Sentence\\in Description} & Uni. 2-gram  \\
\midrule
MiniGPT-4~\cite{zhu2023minigpt} & 483.5 & 5.6 & 84.8 & 45.6 & 4,695 \\
mPLUG-Owl2-7B~\cite{mplug-owl2} & 458.8 & 4.4 & 102.1 & 52.7 & 4,038 \\
LLaVa-1.5-7B~\cite{LLAVA-1.5} & 395.4 & 4.2 & 91.5 & 60.0 & 3,081 \\
InstructBLIP~\cite{instructblip} & 509.8 & 4.0 & 195.4 & 61.3 & 3,260 \\
PALI-5B~\cite{chen2023pali} & 1098.9 & 10.9 & 69.5 & 68.0 & 1,881 \\
VILA~\cite{VILA} & 870.7 & 8.6 & 100.4 & 78.1 & 6,841 \\
mPLUG-Owl3-7B~\cite{ye2024mplugowl3longimagesequenceunderstanding} & 118.0 & 2.0 & 65.2 & 80.4 & 700 \\
LLaVA-Onevision-7B~\cite{llava_onevison} & 672.0 & 6.4 & 107.7 & 81.8 & 5,878 \\
Molmo-7B-D~\cite{MOLMO} & 747.6 & 6.6 & 111.9 & 82.7 & 6,788 \\
LLaVA-Onevision-7B-Chat~\cite{llava_onevison}& 1091.6 & 9.5 & 113.1 & 85.7 & 8,550 \\
Qwen2-VL-7B-Instruct~\cite{Qwen2-VL} & 1022.6 & 9.8 & 102.9 & 87.6 & 8,250 \\
Gemini-1.5-Pro~\cite{geminiteam2024gemini15unlockingmultimodal} & 1326.9 & 12.0 & 109.3 & 95.1 & 11,705 \\
Gemini-1.5-Flash~\cite{geminiteam2024gemini15unlockingmultimodal} & 1199.0 & 11.8 & 100.0 & 96.1 & 10,186 \\
GPT-4o [2024-08-06]~\cite{openai2024gpt4ocard} & 583.5 & 6.2 & 94.2 & 97.1 & 6,160 \\
\midrule
TOTAL & 752.7 & 7.3 & 103.8 & 76.5 & 40,444 \\
\bottomrule
\end{tabular}
}
\label{tab:docci100_details}
\end{table*}

\section{\modelname: Development and Evaluation}
\label{sec:experiments}
\subsection{\modelname{} Model Development}
\label{subsection:model_development}

We developed \textit{\modelname{}} by fine-tuning the 10B parameter PaliGemma-2 architecture~\cite{paligemma2} (details in Appendix) for automated sentence-level factuality assessment and critique generation. This required a specialized training dataset of VLM-generated captions, distinct from \datasetname{}, annotated for factuality and error critiques. To create this diverse training data, we first generated paragraph-length captions using over 70 PaliGemma-2 variants (fine-tuned with varied configurations on DOCCI training data~\cite{DOCCI}) to capture a wide spectrum of generation styles and potential errors. These synthetic captions were then human-annotated per the protocol in \Cref{sec:docci100} (majority vote for labels; longest rationale for non-entailed sentences as critique target).

\modelname{} was fine-tuned on this curated data for two tasks using specific prompts incorporating paragraph context (\texttt{<PREFIX>}Claim-Prefix\texttt{</PREFIX>}), vital for accurate assessment of potentially ambiguous standalone sentences. For \textbf{Factuality Classification}, the prompt was: \textit{``Given the image and the prompt prefix \texttt{<PREFIX>}Claim-Prefix\texttt{</PREFIX>}, does the following text align with the image: \texttt{<TARGET>}Target-Claim\texttt{</TARGET>}?''}, requiring a "Yes"/"No" prediction. For \textbf{Critique Generation}, the prompt was: \textit{``Given the image and the prompt prefix \texttt{<PREFIX>}Claim-Prefix\texttt{</PREFIX>}, the text \texttt{<TARGET>}Target-Claim\texttt{</TARGET>} is considered inaccurate. Explain the misalignments and factual inaccuracies that make it inaccurate.''}. This dual-task strategy enables \modelname{} to identify discrepancies and articulate their reasons.

\subsection{Factuality Classification: Benchmarking and Generalization Results}
\label{subsec:factuality_classification_results} 

This section details the performance of \modelname{} in its factuality classification task, presenting key results from its application as an automated benchmarking tool on \datasetname{} and its generalization capabilities when tested on diverse external datasets.

\paragraph{The \datasetname{} AutoRater: Automated VLM Benchmarking Results.} A primary application of \modelname's classification capability is to serve as an AutoRater for establishing an automated leaderboard that ranks Vision-Language Models (VLMs) based on their factual accuracy when describing images from the \datasetname{} benchmark. The objective is to provide a scalable and reliable alternative to extensive human evaluation for this task. To assess its viability, we evaluated \modelname{} alongside other VLM-based methods as potential automated rankers. We compared how well their automated assessments correlated with human judgments across three distinct factuality criteria: (1) Response-Level Correctness (whether the entire generated paragraph was factually accurate), (2) Percentage of Correct Sentences Overall (total correct sentences across all generated descriptions for a model), and (3) Average Percentage of Correct Sentences per Description. The detailed leaderboards showing the rankings of VLMs for each of these criteria, as determined by both human evaluation and the automated methods, are provided in Appendix. The correlation results, using Spearman's $\rho$ (Sp $\rho$) and Kendall's $\tau$ (Kd $\tau$), are presented in Table~\ref{table:ranking_correlations}. \textbf{\modelname{} demonstrates exceptional performance as an AutoRater, achieving the highest Spearman correlation with human rankings} on Response-Level Correctness (Sp $\rho$ = 0.981) and Percentage of Correct Sentences Overall (Sp $\rho$ = 0.979), and a very high correlation for Average Percentage of Correct Sentences per Description (Sp $\rho$ = 0.968). Its strong performance across these different evaluation granularities, significantly aligning with human assessments, validates its effectiveness as a reliable tool for automatically benchmarking VLM factuality on the \datasetname{} dataset.

\begin{table}[b]

\vspace{-3mm}
\caption{Evaluating Automated Methods as AutoRaters. Correlation (Spearman's $\rho$, Kendall's $\tau$) between model-based rankings and human judgments of VLM factuality on \datasetname{} across three accuracy metrics.
\textbf{Bold} indicates the best score, and \underline{underline} indicates the second best. 
}

\resizebox{\textwidth}{!}{
\begin{tabular}{l cc cc cc}
\toprule
\multicolumn{1}{c}{} &
\multicolumn{2}{c}{\% Response Correct} &
\multicolumn{2}{c}{\% Sentences Overall} &
\multicolumn{2}{c}{\% Sentences per Description} 
\\
\cmidrule(lr{0.5em}){2-3}\cmidrule(lr{0.5em}){4-5}\cmidrule(lr{0.5em}){6-7}
Ranking Method (Model) & Sp $\rho$ & Kd $\tau$ & Sp $\rho$ & Kd $\tau$ & Sp $\rho$ & Kd $\tau$ \\
\midrule

Emu3-Chat & -0.192 & -0.167 & 0.059 & 0.011 & 0.007 & -0.055 \\
InstructBLIP {[}Vicuna-7B{]} & -0.059 & -0.046 & 0.367 & 0.187 & 0.354 & 0.143 \\
Qwen2.5-VL-7B-Instruct & 0.290 & 0.249 & 0.692 & 0.516 & 0.697 & 0.495 \\
Janus-Pro-7B & 0.294 & 0.211 & 0.521 & 0.341 & 0.578 & 0.407 \\
mPLUG-Owl3-7B & 0.734 & 0.573 & 0.741 & 0.582 & 0.798 & 0.648 \\
LLaVa-OneVision{[}Qwen2-7B{]} & 0.889 & 0.760 & 0.855 & 0.758 & 0.851 & 0.736 \\
GPT-4o & 0.920 & 0.818 & 0.975 & 0.911 & \textbf{0.987} & \textbf{0.934} \\
Gemini-2.0-Flash & \underline{0.972} & \underline{0.884} & \underline{0.976} & \underline{0.911} & 0.956 & 0.890 \\
\modelname{} (Ours) & \textbf{0.981} & \textbf{0.928} & \textbf{0.979} & \textbf{0.912} & \underline{0.968} & \underline{0.906} \\
\bottomrule

\end{tabular}

}\label{table:ranking_correlations}

\end{table}

\paragraph{Evaluating Broader Applicability on External Benchmarks.}
To assess \modelname's capabilities beyond our specific benchmark, we evaluate its performance on two established external datasets: M-HalDetect~\cite{mhaldetect}, a benchmark for detecting hallucinations in VLM descriptions of diverse images, and CHOCOLATE~\cite{CHOCOLATE}, which focuses on descriptions of charts and plots. 
We compare \modelname{} against various baselines, including other VLM-based classifiers and embedding-similarity methods, using two standard meta-evaluation metrics: ROC-AUC and Macro-F1. ~\Cref{fig:competitors} provides a qualitative example of these sentence-level classification comparisons.

The performance metrics reported in Table~\ref{tab:claim_classification} were generated based on each model's output type. For models trained to classify via specific output tokens (e.g., `Yes' and `No') – this includes our \modelname{} and other VLM-based classifiers (where scores are derived via a 5-sample strategy) – metrics reflect both confidence and prediction. For all such VLM classifiers, the entailment score for ROC-AUC calculation is obtained by applying a softmax function to the confidence scores associated with the positive and negative classification outputs (yielding a normalized positive probability). The binary classification (`Accurate' or `Inaccurate') for Macro-F1 is determined by selecting the label with the higher confidence score. In contrast, for methods like CLIPScore~\cite{hessel-etal-2021-clipscore}, SigLIP~\cite{SigLIP}, and TIFA~\cite{TIFA}, which output numerical similarity scores, only ROC-AUC is reported, as it directly applies to such scores without requiring an arbitrary threshold.

As shown in Table~\ref{tab:claim_classification}, \textbf{\modelname{} achieves state-of-the-art (SOTA) performance on M-HalDetect. Furthermore, its highly competitive performance on the CHOCOLATE dataset demonstrates notable adaptability and robust reasoning capabilities}, even when evaluating descriptions of charts and graphs without specific training on such visual data. These strong findings across different benchmarks underscore the general utility of our fine-tuned model for factual verification tasks.

\begin{table*}[t]
\caption{Evaluating \modelname{}'s factuality classification: Comparison with baselines on in-distribution (\textsc{\datasetname}) and external (M-HalDetect, CHOCOLATE) datasets. Key results include SOTA on M-HalDetect and strong generalization to CHOCOLATE.}
\resizebox{\textwidth}{!}{  
\begin{tabular}{l|cc|cc|cc}
\toprule
 & \multicolumn{2}{c|}{\textbf{\datasetname}} & \multicolumn{2}{c|}{\textbf{M-HalDetect}} & \multicolumn{2}{c}{\textbf{CHOCOLATE}} \\
\multicolumn{1}{c|}{\textbf{Model}} & ROC-AUC & Macro-F1 & ROC-AUC & Macro-F1 & ROC-AUC & Macro-F1 \\
\midrule
CLIPScore & 0.48 & - & 0.59 & - & 0.56 & - \\
VQAScore {[}CLIP-FlanT5{]} & 0.73 & - & 0.79 & - & 0.71 & - \\
VQAScore {[}GPT-4o{]} & 0.88 & - & 0.85 & - & \textbf{0.84} & - \\
SigLIP & 0.50 & - & 0.63 & - & 0.56 & - \\
TIFA & 0.61 & - & 0.70 & - & 0.57 & - \\
PaliGemma2 {[}9B-448res{]} & 0.51 & 0.23 & 0.61 & 0.39 & 0.53 & 0.00 \\
Qwen2.5-VL-7B-Instruct & 0.65 & 0.36 & 0.81 & 0.75 & 0.81 & 0.74 \\
InstructBLIP {[}Vicuna-7B{]} & 0.50 & 0.45 & 0.45 & 0.40 & 0.53 & 0.37 \\
Emu3-Chat & 0.51 & 0.50 & 0.52 & 0.42 & 0.50 & 0.37 \\
Janus-Pro-7B & 0.67 & 0.58 & 0.72 & 0.59 & 0.65 & 0.47 \\
LLaVa-OneVision{[}Qwen2-7B{]} & 0.76 & 0.58 & 0.82 & 0.60 & 0.75 & 0.44 \\
mPLUG-Owl3-7B & 0.73 & 0.65 & 0.76 & 0.68 & 0.68 & 0.54 \\
Gemini-2.0-Flash & 0.73 & 0.74 & 0.74 & 0.74 & 0.81 & \textbf{0.79} \\
GPT-4o & - & 0.74 & - & 0.69 & - & 0.70 \\
\modelname{} (Ours) & \textbf{0.93} & \textbf{0.83} & \textbf{0.86} & \textbf{0.76} & 0.73 & 0.68 \\
\bottomrule
\end{tabular}
} 

\label{tab:claim_classification}
\end{table*}

\subsection{Evaluating Critique Generation}
\label{subsec:critique_evaluation_results} 
Beyond classifying sentences' correctness, a key capability of \modelname{} is generating textual critiques that explain why a sentence is factually inaccurate. To assess the quality and correctness of these generated explanations, we conducted a dedicated human evaluation study.
The evaluation proceeded as follows: First, we sampled a set of sentences previously identified by human annotators as factually incorrect from both our \datasetname{} benchmark and M-HalDetect dataset. For each sampled incorrect sentence, we prompted \modelname{} as well as several competitive VLMs (listed in \Cref{tab:explanation_correctness_table}) to generate an explanation detailing the specific factual inaccuracies or misalignments, using the critique generation prompt format described in \Cref{subsection:model_development}.
Human annotators were then presented with evaluation instances, each containing: (1) the original image, (2) the specific factually incorrect sentence, and (3) the critique generated by the model being evaluated. The annotators' task was to judge the quality of the critique itself – specifically, whether it accurately and relevantly identified the factual error(s) present in the sentence when compared against the visual evidence in the image. \Cref{tab:explanation_correctness_table} displays the results of this human evaluation, reporting the percentage of generated critiques deemed correct and relevant by the annotators.
The results show that \modelname{} is highly effective at generating useful critiques. On critiques for \datasetname{} sentences, \modelname{} achieves the highest score (73.39\%), slightly surpassing GPT-4o (73.1\%). On critiques for M-HalDetect sentences, \modelname{} performs very strongly (79.33\%), slightly behind Gemini-2.0-Flash (79.89\%) but ahead of GPT-4o (78.77\%). Notably, \modelname{} significantly outperforms several other capable VLMs like Janus-Pro-7B, Qwen-2.5-VL-Instruct, and LLaVA-OV on both datasets. This demonstrates that \modelname{} consistently generates high-quality, accurate explanations for factual errors, a crucial capability for providing interpretable feedback on caption quality and enabling downstream correction tasks.

\begin{table}[h] 
    \caption{Human evaluation of critique quality. Percentage of generated explanations judged as correct and relevant for incorrect sentences sampled from \datasetname{} and M-HalDetect.} 
    \vspace{5pt}

    \centering 
    \label{tab:explanation_correctness_table}
    \begin{tabular}{l|c|c} 
    \toprule
        & \textbf{\datasetname} & \textbf{M-HalDetect} \\ 
        \midrule 
        LLaVA-OV             & 35.96 & 48.04 \\
        Qwen-2.5-VL-Instruct & 45.03 & 58.1  \\ 
        Janus-Pro-7B         & 44.15 & 62.57 \\ 
        Gemini-2.0-Flash     & 64.91 & \textbf{79.89} \\ 
        GPT-4o               & \underline{73.1}  & 78.77 \\ 
        \modelname{} (Ours)                & \textbf{73.39} & \underline{79.33} \\ 
    \bottomrule
    \end{tabular}
\end{table}

\section{Critic-and-Revise}
\label{sec:critic_and_revise}

Many vision-language tasks, from image captioning to text-to-image generation, rely heavily on large-scale datasets of image-text pairs, often utilizing VLM-generated captions for training or as part of their data. For example, large synthetic caption datasets like PixelProse~\cite{singla2024pixelsproselargedataset} are used to train captioning models, and datasets pairing images with descriptive text are fundamental for training text-to-image synthesis models (e.g., leveraging datasets like LAION~\cite{LAION}). However, the factual accuracy and visual alignment of these automatically generated or web-crawled captions can vary, potentially introducing noise or inaccuracies into downstream model training. Improving the quality and factual alignment of such datasets is therefore crucial for advancing these fields. 
Leveraging the critique generation capability of \modelname, we introduce and evaluate a novel \textit{Critic-and-Revise} pipeline. This pipeline is designed not only to correct individual factual inaccuracies in image captions but also offers a pathway to enhance the overall quality of image-text training datasets, thereby potentially improving the performance of models trained on them. This section first outlines the pipeline's methodology (\Cref{subsection:pipeline_methodology}). We then evaluate its applicability in correcting synthetically generated captions from large-scale datasets (\Cref{subsection:correcting_captions}).

\subsection{Pipeline Methodology}
\label{subsection:pipeline_methodology}
The Critic-and-Revise pipeline, illustrated in \Cref{fig:teaser}, operates in two main steps. First, in the \textit{Critic step}, \modelname{} analyzes each sentence of a given caption using its classification function; sentences identified as factually inaccurate trigger the generation of a textual critique explaining the specific error based on the image content. Subsequently, in the \textit{Revise step}, the original inaccurate sentence and its corresponding critique from \modelname{} are used to instruct a separate Large Language Model (LLM) to fix the inaccurate description. For our experiments, we utilized Gemini-2.0-Flash\footnote{Accessed via the Vertex AI API: \url{https://cloud.google.com/vertex-ai}} as the revision LLM. This revision LLM is prompted to rewrite the original sentence, specifically addressing the factual errors highlighted in the critique, while aiming to preserve relevant information and maintain stylistic coherence. The full Critic-and-Revise cycle—factuality classification by \modelname{} for all sentences, followed by critique-guided revision for those flagged as inaccurate—produces a revised caption with enhanced factual alignment to the image.

\subsection{Correcting Synthetically Generated Captions}
\label{subsection:correcting_captions}
To demonstrate the downstream utility of our proposed Critic-and-Revise pipeline, we applied it to captions from two large-scale datasets known for detailed yet potentially unverified descriptions: PixelProse~\cite{singla2024pixelsproselargedataset}, featuring ~16.9M synthetic caption pairs, and DetailCaps-4870~\cite{CAPTURE}, a subset with 4,870 images each accompanied by three detailed synthetic captions. We conducted a human study to evaluate the pipeline's effectiveness: after \modelname{} identified and critiqued inaccurate sentences, and the revision LLM corrected them, human evaluators assessed the factual correctness of both the original flagged sentences (for critic precision) and the revised sentences (for pipeline effectiveness).

The results, summarized in \Cref{tab:critic_revise_combined}, highlight significant improvements. For DetailCaps-4870, while \modelname{}'s initial flagging showed a 15\% false positive rate (original sentences deemed correct by humans), \textbf{the pipeline successfully corrected a large portion of the genuinely inaccurate sentences,} with human judges confirming 61\% of the revised sentences as factually accurate. This represents a 46\% gain in accuracy for the set of sentences initially considered incorrect by the critic. \modelname{}'s own re-evaluation classified 64\% of these revised sentences as accurate, indicating strong self-consistency. Similar positive trends were observed for PixelProse, where human judges found 75\% of revised sentences to be accurate (a 51\% gain), demonstrating the pipeline's capability to enhance factual accuracy in detailed image captions at scale. Qualitative examples illustrating the Critic-and-Revise process, including original incorrect sentences, critiques from \modelname, and the LLM-revised sentences, are provided in Appendix.

\begin{table*}[htbp] 
\centering
\caption{Critic-and-Revise Pipeline factuality: Human and \modelname{} judgments on original vs. revised claims. $\Delta$ = accuracy increase post-revision. \textbf{The pipeline markedly improves claim accuracy (human-confirmed to 61\% on DetailCaps, 75\% on PixelProse for fixed claims, from low initial values)}. \modelname{}'s judgments align, showing high self-consistency.}

\label{tab:critic_revise_combined} 
\begin{tabular}{l ccc @{\hspace{2em}} ccc} 
\toprule
& \multicolumn{3}{c}{\textbf{DetailCaps-4870}} & \multicolumn{3}{c}{\textbf{PixelProse}} \\
\cmidrule(lr){2-4} \cmidrule(lr){5-7} 
\textbf{Judge Type} & \textbf{Original} & \textbf{Fixed} & \textbf{$\Delta$} & \textbf{Original} & \textbf{Fixed} & \textbf{$\Delta$} \\
\midrule
Human Judge        & 15\% & 61\% & +46\% & 24\% & 75\% & +51\% \\
\modelname{} as Judge &  0\% & 64\% & +64\% &  0 & 61\% & +61\% \\ 
\bottomrule
\end{tabular}
\end{table*}

\newpage
\section{Limitations and Future Work}
\label{sec:limitations}
The \datasetname{} benchmark, while richly annotated with 1,400 VLM-generated captions and over 10,000 sentence judgments, is constructed from a base set of 100 unique images. While these images provide diversity and the caption variations are extensive, expanding the number of base images could further enhance the benchmark's statistical power and coverage. Nevertheless, our experiments demonstrate strong generalization capabilities. Specifically, \modelname{}, when trained for factuality classification on \datasetname{}, performs well on external, unseen claim verification datasets like M-HalDetect and CHOCOLATE (\Cref{subsec:factuality_classification_results}). Furthermore, our Critic-and-Revise pipeline, leveraging critiques from \modelname{}, effectively corrects captions on entirely different datasets such as DetailCaps-4870 and PixelProse (\Cref{subsection:correcting_captions}). This collective evidence of generalization across different tasks and datasets suggests the current \datasetname{} size is effective for developing robust and transferable evaluation models and correction methodologies.

Additionally, while \modelname{} achieves strong results in several settings, its performance is not perfect. Our pipeline evaluation (\Cref{sec:critic_and_revise}) indicates that its factuality classification can result in false positives and false negatives (e.g., a 15\% false positive rate on DetailCaps-4870). The quality of its generated critiques, while generally high as shown by human evaluations in \Cref{tab:explanation_correctness_table}, can also exhibit variability. Future work could enhance\modelname{}'s performance by further leveraging our rich annotations. For example, one could investigate methods to merge multiple rationales provided by different annotators into a single, more comprehensive and concise explanation, instead of solely using the longest rationale as the training target for critique generation. Furthermore, our annotation protocol captures whether a sentence's factuality is dependent on the image content or relies on world knowledge (e.g., distinguishing ``The cat is on the mat'' which requires the image, from ``A cat is a mammal'' which does not). This currently unused label could enable a two-stage verification process: first classifying if image grounding is needed, and then applying either the visual fact-checker (\modelname) or a knowledge-based verifier accordingly, potentially improving overall accuracy and efficiency.

Regarding the Critic-and-Revise pipeline, its current design involves two distinct steps: critique generation by \modelname{} followed by revision using a separate LLM. While effective, this contrasts with a hypothetical end-to-end model that might directly output a corrected sentence. However, we argue that the two-step approach offers significant advantages in interpretability. Generating an explicit critique allows for a clear understanding of why a sentence was flagged and what specific error is being addressed. This insight into error types and sources is valuable for analyzing and improving the underlying captioning models, a benefit potentially lost in a direct, black-box correction approach. Therefore, while future work might explore direct revision models, the explanatory power of the intermediate critique remains a key strength of our pipeline.

Addressing these limitations and exploring the suggested avenues for leveraging the full extent of the dataset annotations offers exciting directions for future research in robust and interpretable evaluation of detailed image understanding.

\section{Conclusion}
\label{sec:conclusion}
This work tackled the critical challenge of evaluating and improving the factual accuracy of detailed, paragraph-length VLM-generated image captions. We introduced \textit{\datasetname}, a novel benchmark featuring 1,400 VLM captions with over 10,216 sentence-level human annotations for factuality, including explanatory rationales for errors, providing a vital resource for fine-grained VLM assessment. 
Building on this, we developed \modelname{}, a model proficient in automated factuality classification and critique generation. \modelname{} demonstrated strong generalization with state-of-the-art results on external datasets like M-HalDetect, and its use in the \datasetname{} showed high correlation with human judgments (0.98 Spearman). 
Furthermore, we presented a novel Critic-and-Revise pipeline where \modelname{}'s critiques guide an LLM to automatically correct factual errors, significantly improving caption accuracy as confirmed by human evaluation. 
Collectively, \datasetname, \modelname{}, and the Critic-and-Revise pipeline offer essential tools and methodologies for advancing VLMs towards generating more detailed, fluent, and factually reliable image descriptions. Future work, as outlined in \Cref{sec:limitations}, will explore expanding the benchmark and further enhancing the pipeline's capabilities.


\newpage
{
    \small
\bibliographystyle{ieeenat_fullname}
    \bibliography{main}
}

\newpage
\appendix

\section{\datasetname{} AutoRater Leaderboards}
\label{sec:appendix_autorater_leaderboards}

This appendix presents the complete leaderboards detailing the performance of various Vision-Language Models (VLMs) as automated rankers (AutoRaters) on the \datasetname{} benchmark. These tables supplement the summary correlation metrics (Spearman's $\rho$ and Kendall's $\tau$) found in \Cref{subsec:factuality_classification_results}, \Cref{table:ranking_correlations}. Our goal was to assess how well automated methods, including our \modelname, rank caption-generating VLMs by factual accuracy against human-derived Ground Truth rankings.

Three leaderboards are provided, each for a distinct factuality criterion:
\begin{enumerate}
    \item \textbf{Response-Level Correctness:} Percentage of entirely factually accurate paragraphs (Table~\ref{tab:full_leaderboard_response_correct}).
    \item \textbf{Correct Sentences Overall:} Total percentage of correct sentences across all descriptions (Table~\ref{tab:full_leaderboard_sentence_overall}).
    \item \textbf{Correct Sentences per Description:} Average percentage of correct sentences per description (Table~\ref{tab:full_leaderboard_sentence_average}).
\end{enumerate}

In each leaderboard (Tables~\ref{tab:full_leaderboard_response_correct}, \ref{tab:full_leaderboard_sentence_overall}, and \ref{tab:full_leaderboard_sentence_average}), rows list the 14 caption-generating VLMs from \datasetname{} (details in \Cref{tab:docci100_details}). Columns denote automated ranking methods (e.g., `Ours (\modelname)', `GPT-4o'). Cells show the rank assigned by the column's method to the row's VLM, with the superscript indicating the raw metric score. The final two rows report Spearman's $\rho$ (with p-value superscript) and Kendall's $\tau$ correlations against the Ground Truth for that criterion, offering a nuanced view of each AutoRater's performance.

\begin{table*}[h]
\caption{
VLM AutoRater rankings for Response-Level Correctness on \datasetname. Each cell shows $Rank^{Metric-Score}$. Final two rows: Spearman's $\rho^{p-value}$ and Kendall's $\tau^{p-value}$ correlation against Human.
}
\centering
\label{tab:full_leaderboard_response_correct}
\resizebox{\textwidth}{!}{%

\begin{tabular}{@{}l|c||ccccccccc@{}}
\toprule
\diagbox{Captioner VLM}{Ranking Method} &                       Human &                               Ours  &                         \makecell{Gemini\\2.0-Flash}             & GPT-4o                           & InstructBLIP          & LLaVa-OV                      & Janus-Pro-7B & Qwen2.5-VL & mPLUG-Owl3-7B & Emu3-Chat \\
\midrule

MiniGPT-4               & \rankcell{14}{0.040} & \rankcell{14}{0.060} & \rankcell{14}{0.090} & \rankcell{14}{0.120} & \rankcell{7}{0.960} & \rankcell{14}{0.420} & \rankcell{13}{0.350} & \rankcell{4}{0.0} & \rankcell{13}{0.110} & \rankcell{3}{0.930} \\
MPlugOwl-2              & \rankcell{13}{0.110} & \rankcell{12}{0.120} & \rankcell{13}{0.280} & \rankcell{12}{0.180} & \rankcell{2}{0.980} & \rankcell{12}{0.560} & \rankcell{10}{0.560} & \rankcell{4}{0.0} & \rankcell{12}{0.170} & \rankcell{9}{0.550} \\
LLaVA                   & \rankcell{11}{0.190} & \rankcell{11}{0.170} & \rankcell{9}{0.400} & \rankcell{11}{0.2551} & \rankcell{3}{0.970} & \rankcell{10}{0.710} & \rankcell{5}{0.680} & \rankcell{4}{0.0} & \rankcell{8}{0.210} & \rankcell{12}{0.480} \\
PALI-5B                 & \rankcell{12}{0.112} & \rankcell{13}{0.092} & \rankcell{12}{0.3093} & \rankcell{13}{0.1735} & \rankcell{9}{0.9592} & \rankcell{11}{0.5816} & \rankcell{14}{0.1939} & \rankcell{4}{0.0} & \rankcell{14}{0.0714} & \rankcell{2}{0.9898} \\
VILA                    & \rankcell{10}{0.210} & \rankcell{10}{0.200} & \rankcell{10}{0.390} & \rankcell{9}{0.330} & \rankcell{13}{0.910} & \rankcell{9}{0.850} & \rankcell{1}{0.790} & \rankcell{4}{0.0} & \rankcell{8}{0.210} & \rankcell{4}{0.780} \\
InstructBLIP            & \rankcell{9}{0.260} & \rankcell{9}{0.210} & \rankcell{11}{0.3776} & \rankcell{10}{0.310} & \rankcell{9}{0.9592} & \rankcell{13}{0.540} & \rankcell{9}{0.600} & \rankcell{2}{0.010} & \rankcell{6}{0.260} & \rankcell{10}{0.540} \\
Molmo-7B-D              & \rankcell{8}{0.310} & \rankcell{8}{0.270} & \rankcell{7}{0.680} & \rankcell{8}{0.580} & \rankcell{1}{0.990} & \rankcell{5}{0.930} & \rankcell{4}{0.700} & \rankcell{4}{0.0} & \rankcell{8}{0.210} & \rankcell{4}{0.780} \\
LLaVA-OV-7B-Chat        & \rankcell{7}{0.360} & \rankcell{6}{0.350} & \rankcell{8}{0.630} & \rankcell{5}{0.640} & \rankcell{12}{0.930} & \rankcell{8}{0.900} & \rankcell{8}{0.660} & \rankcell{4}{0.0} & \rankcell{4}{0.300} & \rankcell{8}{0.580} \\
Qwen2-VL-7B-Instruct    & \rankcell{5}{0.410} & \rankcell{4}{0.450} & \rankcell{5}{0.750} & \rankcell{4}{0.650} & \rankcell{3}{0.970} & \rankcell{6}{0.910} & \rankcell{2}{0.780} & \rankcell{4}{0.0} & \rankcell{5}{0.290} & \rankcell{11}{0.490} \\
LLaVA-OV-7B             & \rankcell{5}{0.410} & \rankcell{7}{0.340} & \rankcell{6}{0.720} & \rankcell{7}{0.630} & \rankcell{14}{0.900} & \rankcell{2}{0.970} & \rankcell{6}{0.670} & \rankcell{2}{0.010} & \rankcell{2}{0.410} & \rankcell{7}{0.610} \\
Gemini-1.5-Pro          & \rankcell{4}{0.640} & \rankcell{5}{0.430} & \rankcell{4}{0.840} & \rankcell{3}{0.670} & \rankcell{3}{0.970} & \rankcell{6}{0.910} & \rankcell{12}{0.490} & \rankcell{4}{0.0} & \rankcell{7}{0.240} & \rankcell{14}{0.310} \\
Gemini-1.5-Flash        & \rankcell{3}{0.680} & \rankcell{3}{0.520} & \rankcell{3}{0.880} & \rankcell{5}{0.640} & \rankcell{11}{0.940} & \rankcell{4}{0.940} & \rankcell{11}{0.510} & \rankcell{4}{0.0} & \rankcell{11}{0.180} & \rankcell{13}{0.400} \\
mPLUG-Owl3-7B           & \rankcell{2}{0.710} & \rankcell{2}{0.690} & \rankcell{2}{0.9326} & \rankcell{2}{0.770} & \rankcell{7}{0.960} & \rankcell{1}{0.980} & \rankcell{3}{0.730} & \rankcell{1}{0.020} & \rankcell{1}{0.750} & \rankcell{1}{1.0} \\
GPT-4o[2024-08-06]      & \rankcell{1}{0.830} & \rankcell{1}{0.730} & \rankcell{1}{0.940} & \rankcell{1}{0.890} & \rankcell{3}{0.970} & \rankcell{2}{0.970} & \rankcell{6}{0.670} & \rankcell{4}{0.0} & \rankcell{3}{0.370} & \rankcell{6}{0.740} \\

\midrule 
Spearman's Rank $\rho$ & \multicolumn{1}{c||}{--} & \textbf{0.98}$^{5e-10}$ & \underline{0.97}$^{6e-9}$ & 0.92$^{3e-6}$ & -0.06$^{8e-1}$ & 0.88$^{2e-5}$ & 0.30$^{3e-1}$ & 0.29$^{3e-1}$ & 0.73$^{3e-3}$ & -0.2$^{5e-1}$ \\
Kendall Tau $\tau$ & \multicolumn{1}{c||}{--} & \textbf{0.93}$^{4e-6}$ & \underline{0.89}$^{1e-5}$ & 0.82$^{5e-5}$ & -0.05$^{8e-1}$ & 0.76$^{2e-4}$ & 0.21$^{3e-1}$ & 0.25$^{3e-1}$ & 0.27$^{5e-3}$ & -0.17$^{4e-1}$ \\

\bottomrule

\end{tabular}
}
\end{table*}
\begin{table*}[h]

\caption{
VLM AutoRater rankings for Average Percentage of Correct Sentences on \datasetname. Each cell shows $Rank^{Metric-Score}$. Final two rows: Spearman's $\rho^{p-value}$ and Kendall's $\tau^{p-value}$ correlation against Human.
}

\centering
\label{tab:full_leaderboard_sentence_average}

\resizebox{\textwidth}{!}{%

\begin{tabular}{@{}lc||ccccccccc@{}}
\diagbox{Captioner VLM}{Ranking Method} &                       Human &                 Ours  &                   \makecell{Gemini\\2.0-Flash}     & GPT-4o               & InstructBLIP      & LLaVa-OV          & Janus-Pro-7B           & Qwen2.5-VL & mPLUG-Owl3-7B & Emu3-Chat \\
 \midrule
MiniGPT-4               & \rankcell{14}{0.456} & \rankcell{14}{0.476} & \rankcell{14}{0.529} & \rankcell{14}{0.4242} & \rankcell{8}{0.9931} & \rankcell{14}{0.8385} & \rankcell{13}{0.772} & \rankcell{13}{0.0493} & \rankcell{14}{0.5109} & \rankcell{3}{0.9866} \\
MPlugOwl-2              & \rankcell{13}{0.527} & \rankcell{13}{0.543} & \rankcell{13}{0.6597} & \rankcell{13}{0.5573} & \rankcell{7}{0.9935} & \rankcell{12}{0.8853} & \rankcell{12}{0.8582} & \rankcell{11}{0.0538} & \rankcell{13}{0.5343} & \rankcell{12}{0.8617} \\
LLaVA                   & \rankcell{12}{0.600} & \rankcell{11}{0.589} & \rankcell{11}{0.7475} & \rankcell{12}{0.5876} & \rankcell{9}{0.9917} & \rankcell{11}{0.913} & \rankcell{9}{0.9018} & \rankcell{9}{0.0817} & \rankcell{11}{0.5798} & \rankcell{14}{0.8325} \\
InstructBLIP            & \rankcell{11}{0.613} & \rankcell{12}{0.581} & \rankcell{12}{0.7054} & \rankcell{11}{0.6225} & \rankcell{12}{0.9804} & \rankcell{13}{0.8619} & \rankcell{10}{0.8902} & \rankcell{10}{0.0544} & \rankcell{12}{0.5638} & \rankcell{13}{0.840} \\
PALI-5B                 & \rankcell{10}{0.672} & \rankcell{10}{0.675} & \rankcell{10}{0.7948} & \rankcell{10}{0.6711} & \rankcell{3}{0.9966} & \rankcell{10}{0.9133} & \rankcell{14}{0.7338} & \rankcell{12}{0.0518} & \rankcell{10}{0.6068} & \rankcell{2}{0.9992} \\
VILA                    & \rankcell{9}{0.781} & \rankcell{8}{0.784} & \rankcell{9}{0.8604} & \rankcell{9}{0.8062} & \rankcell{11}{0.9896} & \rankcell{9}{0.9832} & \rankcell{1}{0.9697} & \rankcell{4}{0.2229} & \rankcell{8}{0.7644} & \rankcell{4}{0.9619} \\
mPLUG-Owl3-7B           & \rankcell{8}{0.804} & \rankcell{6}{0.799} & \rankcell{4}{0.9781} & \rankcell{8}{0.8675} & \rankcell{13}{0.9767} & \rankcell{5}{0.9921} & \rankcell{11}{0.8655} & \rankcell{14}{0.0475} & \rankcell{2}{0.8474} & \rankcell{1}{1.0} \\
LLaVA-OV-7B             & \rankcell{7}{0.818} & \rankcell{6}{0.799} & \rankcell{7}{0.9433} & \rankcell{6}{0.9052} & \rankcell{14}{0.9736} & \rankcell{1}{0.9962} & \rankcell{5}{0.9366} & \rankcell{6}{0.1959} & \rankcell{7}{0.8102} & \rankcell{8}{0.9189} \\
Molmo-7B-D              & \rankcell{6}{0.827} & \rankcell{9}{0.783} & \rankcell{6}{0.9457} & \rankcell{7}{0.9015} & \rankcell{1}{0.9983} & \rankcell{7}{0.9895} & \rankcell{4}{0.9385} & \rankcell{8}{0.0987} & \rankcell{9}{0.725} & \rankcell{5}{0.953} \\
LLaVA-OV-7B-Chat        & \rankcell{5}{0.857} & \rankcell{5}{0.854} & \rankcell{8}{0.9415} & \rankcell{4}{0.9364} & \rankcell{10}{0.9912} & \rankcell{8}{0.9883} & \rankcell{3}{0.944} & \rankcell{1}{0.2779} & \rankcell{1}{0.8482} & \rankcell{7}{0.9305} \\
Qwen2-VL-7B-Instruct    & \rankcell{4}{0.876} & \rankcell{4}{0.884} & \rankcell{5}{0.9677} & \rankcell{5}{0.934} & \rankcell{4}{0.9962} & \rankcell{6}{0.9915} & \rankcell{2}{0.9681} & \rankcell{5}{0.2227} & \rankcell{5}{0.8275} & \rankcell{9}{0.9078} \\
Gemini-1.5-Pro          & \rankcell{3}{0.951} & \rankcell{3}{0.925} & \rankcell{3}{0.9856} & \rankcell{2}{0.9664} & \rankcell{2}{0.9968} & \rankcell{4}{0.9928} & \rankcell{8}{0.9282} & \rankcell{3}{0.251} & \rankcell{4}{0.8341} & \rankcell{11}{0.8929} \\
Gemini-1.5-Flash        & \rankcell{2}{0.961} & \rankcell{2}{0.939} & \rankcell{2}{0.9894} & \rankcell{3}{0.9544} & \rankcell{5}{0.9942} & \rankcell{3}{0.9948} & \rankcell{7}{0.9342} & \rankcell{2}{0.271} & \rankcell{3}{0.8381} & \rankcell{10}{0.9017} \\
GPT-4o[2024-08-06]      & \rankcell{1}{0.971} & \rankcell{1}{0.949} & \rankcell{1}{0.9914} & \rankcell{1}{0.9801} & \rankcell{6}{0.9942} & \rankcell{2}{0.995} & \rankcell{6}{0.9345} & \rankcell{7}{0.1664} & \rankcell{6}{0.8233} & \rankcell{6}{0.9477} \\
\midrule 
Spearman's Rank $\rho$ & \multicolumn{1}{c||}{--}   & \underline{0.97}$^{1e-8}$ & 0.96$^{9e-8}$ & \textbf{0.99}$^{7e-11}$ & 0.35$^{2e-1}$ & 0.85$^{1e-4}$ & 0.58$^{3e-2}$ & 0.70$^{5e-3}$ & 0.80$^{6e-4}$ & 0.00$^{1e-0}$ \\
Kendall Tau $\tau$ & \multicolumn{1}{c||}{--}       & \underline{0.91}$^{7e-6}$ & 0.90$^{2e-7}$  & \textbf{0.93}$^{1e-8}$ & 0.14$^{5e-1}$ & 0.74$^{7e-5}$ & 0.40$^{4e-2}$ & 0.50$^{1e-2}$ & 0.65$^{7e-4}$ & -0.05$^{8e-1}$ \\
\bottomrule
\end{tabular}
}
\end{table*} 
\begin{table*}[h]
\caption{
VLM AutoRater rankings for Percentage of Correct Sentences Overall on \datasetname. Each cell shows $Rank^{Metric-Score}$. Final two rows: Spearman's $\rho^{p-value}$ and Kendall's $\tau^{p-value}$ correlation against Human.
}
\centering
\label{tab:full_leaderboard_sentence_overall}
\resizebox{\textwidth}{!}{%

\begin{tabular}{@{}lc||ccccccccc@{}}
\toprule
\diagbox{Captioner VLM}{Ranking Method} &                       Human &                               Ours  &                         \makecell{Gemini\\2.0-Flash}             & GPT-4o                           & InstructBLIP          & LLaVa-OV                      & Janus-Pro-7B & Qwen2.5-VL & mPLUG-Owl3-7B & Emu3-Chat \\
\midrule
MiniGPT-4               & \rankcell{14}{0.477} & \rankcell{14}{0.493} & \rankcell{14}{0.550} & \rankcell{14}{0.444} & \rankcell{7}{0.993} & \rankcell{14}{0.825} & \rankcell{13}{0.768} & \rankcell{11}{0.055} & \rankcell{13}{0.518} & \rankcell{3}{0.988} \\
MPlugOwl-2              & \rankcell{13}{0.516} & \rankcell{13}{0.532} & \rankcell{13}{0.642} & \rankcell{13}{0.562} & \rankcell{7}{0.993} & \rankcell{12}{0.872} & \rankcell{11}{0.847} & \rankcell{13}{0.052} & \rankcell{14}{0.514} & \rankcell{12}{0.863} \\
InstructBLIP            & \rankcell{12}{0.574} & \rankcell{12}{0.567} & \rankcell{12}{0.677} & \rankcell{11}{0.588} & \rankcell{11}{0.990} & \rankcell{13}{0.839} & \rankcell{10}{0.869} & \rankcell{12}{0.054} & \rankcell{12}{0.537} & \rankcell{14}{0.824} \\

LLaVA                   & \rankcell{11}{0.594} & \rankcell{11}{0.585} & \rankcell{10}{0.741} & \rankcell{12}{0.587} & \rankcell{7}{0.993} & \rankcell{10}{0.903} & \rankcell{9}{0.894} & \rankcell{9}{0.083} & \rankcell{11}{0.583} & \rankcell{13}{0.828} \\
PALI-5B                 & \rankcell{10}{0.671} & \rankcell{10}{0.663} & \rankcell{10}{0.741} & \rankcell{10}{0.620} & \rankcell{4}{0.996} & \rankcell{11}{0.877} & \rankcell{14}{0.725} & \rankcell{10}{0.067} & \rankcell{10}{0.591} & \rankcell{2}{0.999} \\
VILA                    & \rankcell{9}{0.789} & \rankcell{8}{0.790} & \rankcell{9}{0.859} & \rankcell{9}{0.805} & \rankcell{11}{0.990} & \rankcell{9}{0.983} & \rankcell{1}{0.969} & \rankcell{4}{0.224} & \rankcell{8}{0.769} & \rankcell{4}{0.962} \\
LLaVA-OV-7B             & \rankcell{8}{0.815} & \rankcell{7}{0.812} & \rankcell{8}{0.945} & \rankcell{6}{0.906} & \rankcell{13}{0.984} & \rankcell{1}{0.995} & \rankcell{5}{0.934} & \rankcell{6}{0.213} & \rankcell{2}{0.845} & \rankcell{9}{0.906} \\
mPLUG-Owl3-7B           & \rankcell{7}{0.822} & \rankcell{6}{0.822} & \rankcell{4}{0.960} & \rankcell{8}{0.840} & \rankcell{14}{0.980} & \rankcell{8}{0.985} & \rankcell{12}{0.812} & \rankcell{14}{0.051} & \rankcell{3}{0.838} & \rankcell{1}{1.0} \\
Molmo-7B-D              & \rankcell{6}{0.831} & \rankcell{9}{0.786} & \rankcell{6}{0.946} & \rankcell{7}{0.903} & \rankcell{1}{0.998} & \rankcell{6}{0.989} & \rankcell{4}{0.937} & \rankcell{8}{0.098} & \rankcell{9}{0.718} & \rankcell{5}{0.955} \\
Qwen2-VL-7B-Instruct    & \rankcell{5}{0.867} & \rankcell{4}{0.877} & \rankcell{4}{0.960} & \rankcell{5}{0.929} & \rankcell{2}{0.997} & \rankcell{5}{0.991} & \rankcell{2}{0.966} & \rankcell{5}{0.219} & \rankcell{6}{0.833} & \rankcell{8}{0.907} \\
LLaVA-OV-7B-Chat        & \rankcell{4}{0.869} & \rankcell{5}{0.875} & \rankcell{6}{0.946} & \rankcell{4}{0.935} & \rankcell{10}{0.991} & \rankcell{7}{0.988} & \rankcell{3}{0.958} & \rankcell{1}{0.311} & \rankcell{1}{0.868} & \rankcell{7}{0.924} \\
Gemini-1.5-Pro          & \rankcell{3}{0.954} & \rankcell{3}{0.925} & \rankcell{3}{0.985} & \rankcell{2}{0.968} & \rankcell{2}{0.997} & \rankcell{4}{0.992} & \rankcell{8}{0.929} & \rankcell{3}{0.250} & \rankcell{5}{0.835} & \rankcell{11}{0.891} \\
Gemini-1.5-Flash        & \rankcell{2}{0.962} & \rankcell{2}{0.937} & \rankcell{1}{0.990} & \rankcell{3}{0.953} & \rankcell{5}{0.995} & \rankcell{1}{0.995} & \rankcell{5}{0.934} & \rankcell{2}{0.269} & \rankcell{4}{0.836} & \rankcell{9}{0.906} \\
GPT-4o[2024-08-06]      & \rankcell{1}{0.969} & \rankcell{1}{0.950} & \rankcell{1}{0.990} & \rankcell{1}{0.979} & \rankcell{5}{0.995} & \rankcell{1}{0.995} & \rankcell{5}{0.934} & \rankcell{7}{0.161} & \rankcell{7}{0.808} & \rankcell{6}{0.947} \\
\midrule 
Spearman's Rank $\rho$ & \multicolumn{1}{c||}{--} & \textbf{0.98}$^{1e-9}$ & \underline{0.98}$^{2e-9}$ & 0.97$^{2e-9}$ & 0.37$^{2e-1}$ & 0.86$^{10e-5}$ & 0.52$^{6e-2}$ & 0.69$^{6e-3}$ & 0.74$^{2e-3}$ & 0.06$^{8e-1}$ \\
Kendall Tau $\tau$ & \multicolumn{1}{c||}{--} & \textbf{0.91}$^{5e-8}$ & \underline{0.91}$^{5e-8}$ & 0.91$^{5e-8}$ & 0.19$^{4e-1}$ & 0.76$^{4e-5}$ & 0.34$^{1e-1}$ & 0.52$^{10e-3}$ & 0.58$^{3e-3}$ & 0.01$^{1e+0}$ \\

\bottomrule

\end{tabular}
}
\end{table*}       

\section{\modelname{} Model Development Details}
\label{sec:appendix_vnli_critique_dev}

This section provides further details on the architecture, fine-tuning process, and computational resources utilized for the development of our \modelname{} model, as introduced in \Cref{sec:experiments} of the main paper.

\subsection{Model Architecture}
\label{subsec:appendix_model_architecture}

VNLI-Critique is developed by fine-tuning the PaliGemma 10B architecture \cite{paligemma2}. This architecture integrates a Gemma2-9B Large Language Model (LLM)~\cite{GEMMA2} as its textual backbone and a SigLIP model~\cite{SigLIP} as its visual encoder. For visual processing, input images are standardized to a resolution of $448px^2$ pixels. At this resolution, the SigLIP visual encoder processes each image into a sequence of 1024 visual tokens, which are subsequently fed into the LLM component for multimodal understanding and generation tasks.

\subsection{Fine-tuning Procedure}
\label{subsec:appendix_finetuning_procedure}

We performed full fine-tuning of the PaliGemma 10B model to develop VNLI-Critique. The fine-tuning process was conducted for 5 epochs. A batch size of 128 was used, with a dropout rate of 0.1 applied to aid regularization. No weight decay was utilized during training. The Adam optimizer \cite{ADAM_OPTIMIZER} was employed with its default hyperparameters, and a constant learning rate of $1 \times 10^{-6}$ was maintained throughout the fine-tuning process.

\subsection{Computational Resources}
\label{subsec:appendix_computational_resources}

The training of the \modelname{} model was executed on Google Cloud TPUv5e~\cite{tpucloud} accelerators. Specifically, a configuration of 128 TPUv5e chips was utilized for the fine-tuning task. The total training time for the 5 epochs was approximately 1 hour and 30 minutes. Based on an estimated cost of \$1.20 per chip-hour, the total computational cost for training VNLI-Critique was approximately \$230.40.

\section{Human Annotation Details}
\label{sec:appendix_human_annotation}

The creation of the \datasetname{} benchmark and the evaluation of our models' outputs, including critique generation and the Critic-and-Revise pipeline, relied on comprehensive human annotations. We engaged third-party human annotators sourced through Prolific\footnote{\url{https://www.prolific.com/}}. Each data entry subject to human evaluation, whether for sentence-level factuality in \datasetname{} or for the quality assessment of generated critiques, was independently assessed by five different annotators. This multi-annotator approach helps ensure robustness and mitigate individual biases in the collected judgments. Annotators were compensated at a rate of \$20 per hour for their work.

The following subsections provide an illustrative overview of the annotation interfaces designed for the two primary human evaluation tasks: assessing the factuality of VLM-generated description sentences (Section~\ref{subsec:appendix_desc_sentence_interface}) and evaluating the quality of generated critiques (Section~\ref{subsec:appendix_critique_interface}).

\subsection{Description Sentences Annotation Interface}
\label{subsec:appendix_desc_sentence_interface}

\begin{figure*}[h]
  \centering
  \includegraphics[width=1\textwidth]{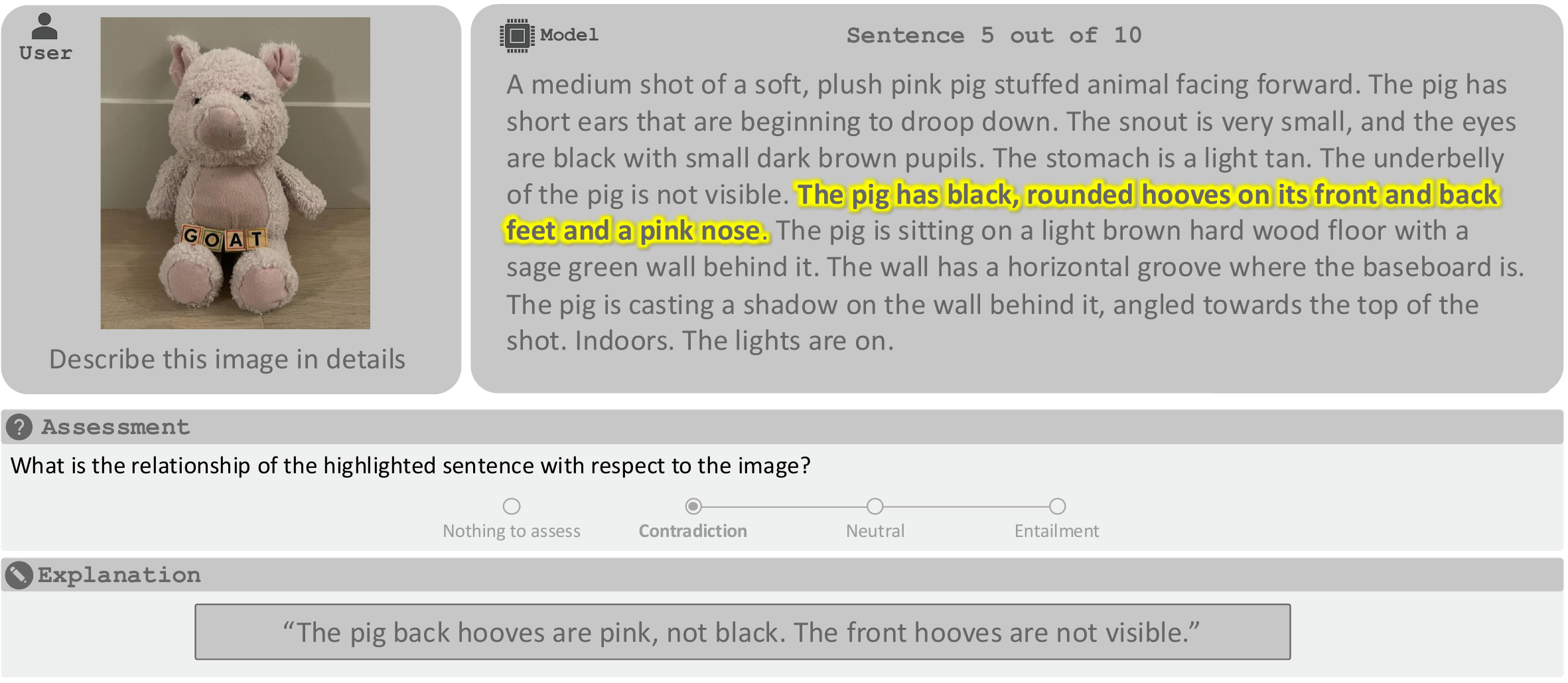}

\caption{Example of the Description Sentences Annotation Interface. Annotators are shown the image, the full VLM-generated paragraph, and a highlighted sentence. They assess its factuality by selecting a label (here, `Contradiction') and providing a textual explanation for any inaccuracies observed.}
\label{fig:sentence_annotation_template}

\end{figure*}

For the task of annotating sentence-level factuality within VLM-generated paragraph descriptions (as detailed in Section 3 for the \datasetname{} benchmark), annotators were presented with an interface displaying the source image, the full paragraph context, and the specific sentence under evaluation. Figure~\ref{fig:sentence_annotation_template} illustrates a representative example of this annotation interface. Annotators were asked to judge whether the sentence accurately described the image content, providing labels such as 'Entailment', 'Neutral', or 'Contradiction', and to supply textual rationales for any non-entailed judgments.

\subsection{Critique Annotation Interface}
\label{subsec:appendix_critique_interface}

\begin{figure*}
  \centering
  \includegraphics[width=1\columnwidth]{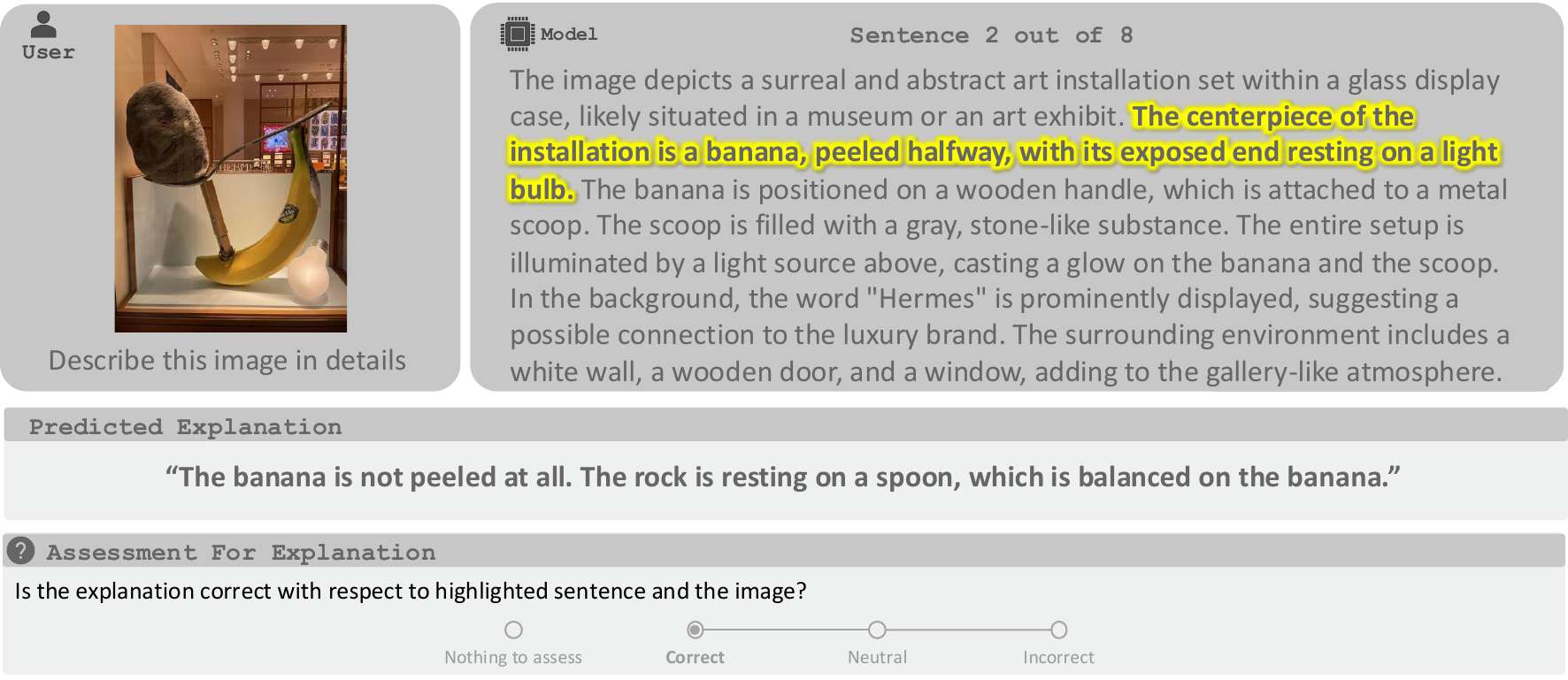}

\caption{Example of the Critique Annotation Interface. Annotators assess if the 'Predicted Explanation' correctly identifies the error in the VLM's highlighted sentence relative to the image.}
\label{fig:explanation_annotation_template}

\end{figure*}

To evaluate the quality of critiques generated by VNLI-Critique and other baseline models (as described in Section 4.3), a different interface was employed. This interface presented human annotators with the original image, the factually incorrect sentence that was critiqued, and the critique generated by the model under evaluation. Figure~\ref{fig:explanation_annotation_template}  shows an example of this interface. Annotators were tasked with judging whether the provided critique accurately and relevantly identified the factual error(s) present in the original sentence when compared against the visual evidence in the image.

\section{Qualitative Examples}
\label{sec:appendix_qualitative_examples}

To further illustrate the core components and outputs of our work, this section provides additional qualitative examples, complementing the discussions and aggregated results presented in the main paper.

Table~\ref{table:appendix_annotation_example} showcases another detailed entry from the \datasetname{} benchmark. This example highlights the fine-grained nature of our sentence-level annotations, including the multi-rater judgments on whether a sentence makes a claim about the image, its factual correctness against the visual evidence, and the diverse human-written rationales provided by annotators for any identified inaccuracies. Such examples underscore the richness of the benchmark for evaluating nuanced understanding and error analysis.

Furthermore, Table~\ref{table:critic_and_revise_example} provides a step-by-step walkthrough of our Critic-and-Revise pipeline operating on an image description sourced from the PixelProse~\cite{singla2024pixelsproselargedataset} dataset. The example demonstrates: (1) the original VLM-generated description containing factual errors, (2) the specific unfactual sentences detected by \modelname, (3) the corresponding critiques generated by \modelname, (4) the individual sentence revisions made by the LLM based on these critiques, and (5) the final, more factually accurate revised description. This illustrates the practical application of our pipeline in automatically correcting errors in detailed image captions.
\begin{table*}

\caption{Additional \datasetname{} benchmark annotation example (5 raters per assessment). Details sentence-level claims, factuality, and diverse human rationales for errors, showing varied perspectives.}
\centering
\scriptsize 
\setlength{\tabcolsep}{3pt} 
\renewcommand{\arraystretch}{0.9} 
\renewcommand\tabularxcolumn[1]{m{#1}}

\begin{tabularx}{\textwidth}{@{} *{4}{>{\centering\arraybackslash}X} @{}} 

\toprule

{\textbf{Image}} & \multicolumn{2}{c}{\raisebox{-0.5\height}{\includegraphics[width=0.25\textwidth]{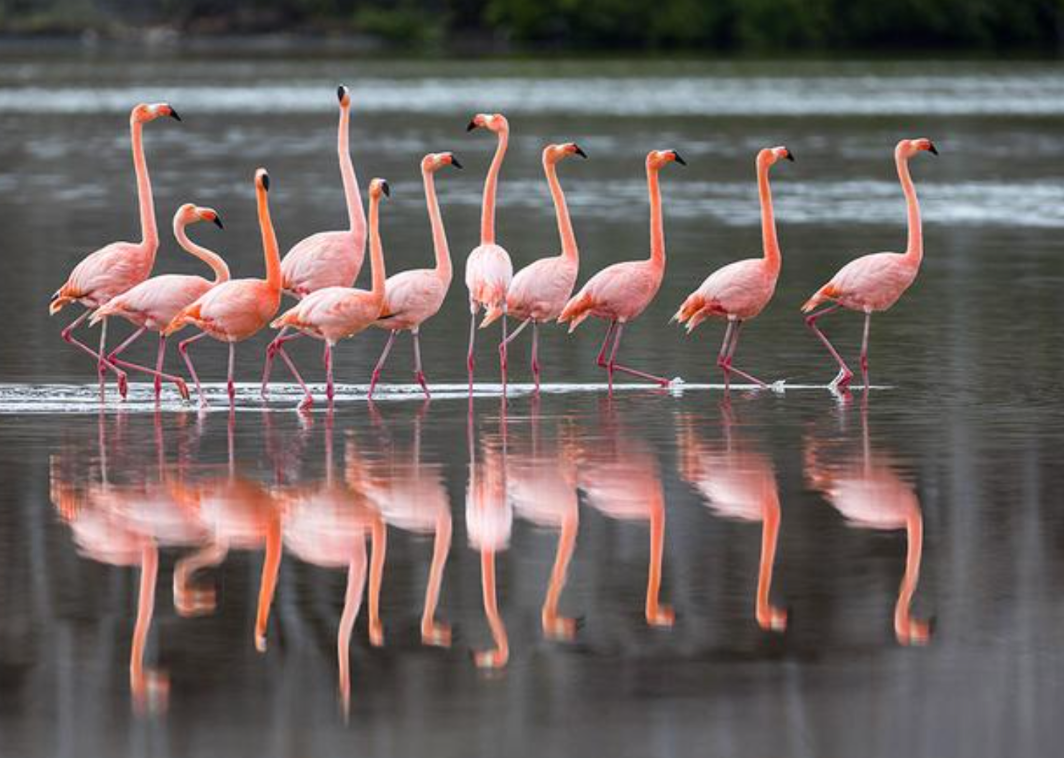}}} \\

\midrule
\textbf{\makecell{Description \\ Sentence}} & ``... Looking closely, we can see eight flamingos lined up. ...'' & ``... They are standing in a body of water, their reflection is seen in the water, and there are trees in the background. ...'' & ``... Flamingos primarily eat brine shrimp, blue-green algae, small insects, mollusks, and crustaceans ...'' \\
\midrule
\textbf{Does the sentence include a claim about the image? (Answers from 5 raters)} 
    & \checkemoji, \checkemoji, \checkemoji, \checkemoji, \checkemoji 
    & \checkemoji, \checkemoji, \checkemoji, \checkemoji, \checkemoji 
    & \crossemoji, \crossemoji, \crossemoji, \crossemoji, \crossemoji \\
\midrule
\textbf{Is the sentence factual? (Answers from 5 raters)} 
    & \crossemoji, \crossemoji, \crossemoji, \crossemoji, \crossemoji
    & \crossemoji, \checkemoji, \checkemoji, \crossemoji, \checkemoji 
    & \checkemoji, \checkemoji, \checkemoji, \checkemoji, \checkemoji \\
\midrule

\textbf{Rationales} & \begin{itemize}[leftmargin=*] 
    \item The count of eight flamingos is incorrect; I can see at least ten.
    \item There are 11 flamingos in the image. 
    \item Incorrect number of flamingos stated, there appear to be more.
    \item I see eleven flamingos, not eight.
    \item Eleven flamingos are lined up, not eight.
\end{itemize} & \begin{itemize}[leftmargin=*]
    \item I don't see any prominent trees in the background, mostly just distant, blurry foliage or land.
    \item The background appears to be more of a distant shoreline or low vegetation, not distinct trees.
\end{itemize} & - \\
\bottomrule
\end{tabularx}
\label{table:appendix_annotation_example}
\vspace{-15pt}
\end{table*}

\begin{table*}[h]
\caption{Table 11: Step-by-step illustration of the \textbf{Critic-and-Revise} pipeline in action on a sample from the PixelProse dataset. The `Original Description' contains several inaccuracies. `Detected Unfactual Sentences by \modelname' highlights these errors (e.g., regarding hand position, light source, text location). `Predicted Critiques by \modelname' provides explanations for these errors. The `Critic-and-Revise output' shows the LLM's corrected individual sentences, guided by the critiques. Finally, the 'Revised Description' integrates these corrections into a more factually grounded paragraph.}
\centering
\scriptsize
\setlength{\tabcolsep}{3pt}
\renewcommand{\arraystretch}{0.9}
\renewcommand\tabularxcolumn[1]{m{#1}} 

\newlength{\myDescWidth}
\setlength{\myDescWidth}{\dimexpr (3\textwidth - 2\tabcolsep - 4\tabcolsep)/4*3 + 4\tabcolsep \relax} 
\setlength{\myDescWidth}{0.75\textwidth - 4\tabcolsep} 

\setlength{\myDescWidth}{\dimexpr (3\textwidth - 2\tabcolsep)/4 \relax} 

\begin{tabularx}{\textwidth}{@{} *{4}{>{\centering\arraybackslash}X} @{}} 
\toprule

\textbf{Image} &
\multicolumn{2}{c}{\raisebox{-0.5\height}{\includegraphics[width=0.25\textwidth]{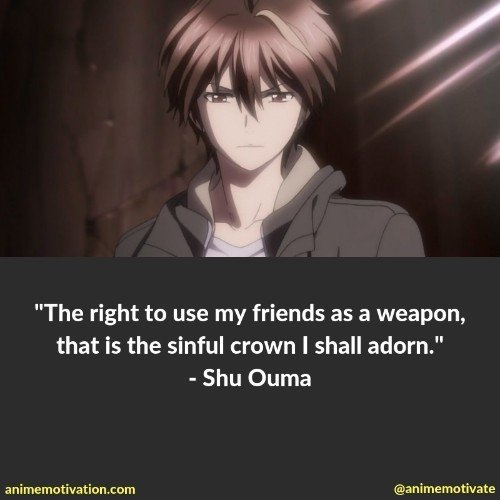}}} & 
\\

\midrule

\raisebox{-3.5\height}{\textbf{Original Description}} &
\multicolumn{3}{p{\myDescWidth}}{\raggedright 
A young man with short brown hair and dark brown eyes.
He is wearing a black jacket and a white shirt.
He has a serious expression on his face.
He is looking at the viewer with his left hand on his chin and the other holding his jacket.
There is a dark background with some light coming from the left side of the image.
There is text at the top of the image that says ``The right to use my friends as a weapon, that is the sinful crown I shall adorn - Shu Ouma''.
The text is in a white font. The image is in an anime style.
} \\

\midrule

\textbf{Detected Unfactual Sentences by \textit{\modelname}} &
He is looking at the viewer {\color{red}with his left hand on his chin and the other holding his jacket.}  &
There is a dark background with some light coming from the {\color{red}left} side of the image.  &
There is text at the {\color{red}top} of the image that says ``The right to use my friends as a weapon, that is the sinful crown I shall adorn - Shu Ouma''. \\
\midrule

\textbf{Predicted Critiques by \textit{\modelname}} &
He is looking at the viewer but his hands are not visible &
The light is coming from the right side of the image, not the left. &
The text is at the bottom of the image and not the top. \\ 
\midrule

\textbf{Critic-and-Revise output} &
He is looking at the viewer. &
There is a dark background with some light coming from the {\color{green}right} side of the image. &
There is text at the {\color{green}bottom} of the image that says ``The right to use my friends as a weapon, that is the sinful crown I shall adorn - Shu Ouma''. \\ 
\midrule

\raisebox{-3.5\height}{\textbf{Revised Description}} &
\multicolumn{3}{p{\myDescWidth}}{\raggedright 
A young man with short brown hair and dark brown eyes.
He is wearing a black jacket and a white shirt.
He has a serious expression on his face.
He is looking at the viewer.
There is a dark background with some light coming from the right side of the image.
There is text at the bottom of the image that says ``The right to use my friends as a weapon, that is the sinful crown I shall adorn - Shu Ouma''.
The text is in a white font. The image is in an anime style.
} \\

\bottomrule
\end{tabularx}
\label{table:critic_and_revise_example}
\end{table*}

\newpage
\clearpage


\end{document}